\documentclass{article} 
\usepackage{iclr2019_conference}

\usepackage[title]{appendix}

\usepackage{times}
\usepackage{latexsym}

\usepackage{url}

\usepackage{amsmath}
\usepackage{bbm}

\usepackage{amssymb}
\usepackage{algorithm}
\usepackage{algpseudocode}
\usepackage{multirow}
\usepackage{tabu}

\usepackage{graphicx} 
\usepackage{subfigure}
\usepackage{booktabs}       
\usepackage{breakcites}

\usepackage{array}
\newcolumntype{H}{>{\setbox0=\hbox\bgroup}c<{\egroup}@{}}

\newcommand{\RR}[0]{\mathbb{R}}

\iclrfinalcopy

\title{Learning to Coordinate Multiple Reinforcement Learning Agents
for Diverse Query Reformulation}

\author{Rodrigo Nogueira\thanks{Work done while interning at Google.} \\
New York University\\
\texttt{rodrigonogueira@nyu.edu} \\
\AND
Jannis Bulian, Massimiliano Ciaramita \\
Google AI Language\\
\texttt{\{jbulian,massi\}@google.com} \\
}

\begin{document}
\maketitle
\begin{abstract}
We propose a method to efficiently learn diverse strategies in
reinforcement learning for query reformulation in the tasks of document retrieval
and question answering. In the proposed framework an agent
consists of multiple specialized sub-agents and a meta-agent that
learns to aggregate the answers from sub-agents to produce a final
answer. Sub-agents are trained on disjoint partitions of the training
data, while the meta-agent is trained on the full training set. Our
method makes learning faster, because it is highly parallelizable, and
has better generalization performance than strong baselines, such as
an ensemble of agents trained on the full data. We show that the
improved performance is due to the increased diversity of
reformulation strategies. 
\end{abstract}

\section{Introduction}

Reinforcement learning has proven effective in several language
processing tasks, such as machine
translation~\citep{wu2016google,ranzato2015sequence,bahdanau2016actor},
question-answering~\citep{wang2017r,hu2017reinforced}, and text
summarization~\citep{paulus2017deep}. 
In reinforcement learning efficient exploration is key to achieve good
performance.
The ability to explore in parallel a diverse set of
strategies often speeds up training and leads to a better
policy~\citep{mnih2016asynchronous,osband2016deep}. 

In this work, we propose a simple method to achieve efficient
parallelized exploration of diverse policies, inspired by hierarchical
reinforcement learning
\citep{singh1992reinforcement,lin1993reinforcement,
  dietterich2000hierarchical, dayan1993feudal}. 
We structure the agent into multiple
\textit{sub-agents}, which are trained on disjoint subsets of the
training data. Sub-agents are co-ordinated by a meta-agent,
called \textit{aggregator}, that groups and scores answers from the
sub-agents for each given input. Unlike sub-agents, the aggregator is a
generalist since it learns a policy for the entire training set. 

We argue that it is easier to train multiple sub-agents than a single
generalist one since each sub-agent only needs to learn a policy that
performs well for a subset of examples. 
Moreover, specializing agents on different partitions of the data
encourages them to learn distinct policies, thus giving the aggregator
the possibility to see answers from a population of diverse agents.
Learning a single policy that results in an equally diverse strategy is more challenging.  

Since each sub-agent is trained on a fraction of the data, and there is no
communication between them, training can be done faster than training
a single agent on the full data. Additionally, it is easier to
parallelize than applying existing distributed algorithms such as
asynchronous SGD or A3C~\citep{mnih2016asynchronous}, as the sub-agents
do not need to exchange weights or gradients. After training the
sub-agents, only their actions need to be sent to the aggregator.


We build upon the works of \citet{nogueira2017task} and
\citet{aqa-iclr:2018}. Therefore, we evaluate the proposed method on the
same tasks they used: query reformulation for document retrieval and
question-answering. We show that it outperforms a strong baseline of
an ensemble of agents trained on the full dataset. We also found that
performance and reformulation diversity are correlated (Sec.~\ref{sec:query_diversity}).  

Our main contributions are the following:
\begin{itemize}
\setlength\itemsep{1pt}
\item A simple method to achieve more diverse strategies and better generalization performance than a model average ensemble.
\item Training can be easily parallelized in the proposed method.
\item An interesting finding that contradicts our, perhaps naive,
  intuition: specializing agents on semantically similar data does not
  work as well as random partitioning. An explanation is given in Appendix~\ref{sec:partitioning_methods}. 
\end{itemize}

\section{Related Work}

The proposed approach is inspired by the mixture of experts, which was
introduced more than two decades
ago~\citep{jacobs1991adaptive,jordan1994hierarchical} and has been a
topic of intense study since then. The idea consists of training a
set of agents, each specializing in some task or data. One or
more gating mechanisms then select subsets of the agents that will
handle a new input. Recently, \citet{shazeer2017outrageously}
revisited the idea and showed strong performances in the supervised
learning tasks of language modeling and machine translation. Their
method requires that output vectors of experts are exchanged between
machines. Since these vectors can be large, the network bandwidth
becomes a bottleneck. They used a variety of techniques to mitigate
this problem.~\citet{anil2018large} later proposed a method to further
reduce communication overhead by only exchanging the probability distributions of the different agents. Our method, instead, requires only scalars (rewards) and short strings (original query, reformulations, and answers) to be exchanged. Therefore, the communication overhead is small. 

Previous works used specialized agents to improve exploration in
RL~\citep{dayan1993feudal,singh1992reinforcement,kaelbling1996reinforcement}. For
instance, \citet{stanton2016curiosity} and \citet{conti2017improving}
use a population of agents to achieve a high diversity of strategies
that leads to better generalization performance and faster
convergence. ~\citet{rusu2015policy} use experts to learn subtasks and
later merge them into a single agent using
distillation~\citep{hinton2015distilling}. 

The experiments are often carried out in simulated environments, such
as robot control~\citep{brockman2016openai} and
video-games~\citep{bellemare2013arcade}. In these environments, rewards
are frequently available, the states have low diversity (e.g., same
image background), and responses usually are fast (60 frames per
second). We, instead, evaluate our approach on tasks whose inputs
(queries) and states (documents and answers) are diverse because they
are in natural language, and the environment responses are slow (0.5-5
seconds per query). 

Somewhat similarly motivated is the work
of~\citet{serban2017deep}. They train many heterogeneous response
models and further train an RL agent to pick one response per
utterance.

\section{Method}
\begin{figure*}[t]
\begin{center}
\centerline{\includegraphics[width=\textwidth]{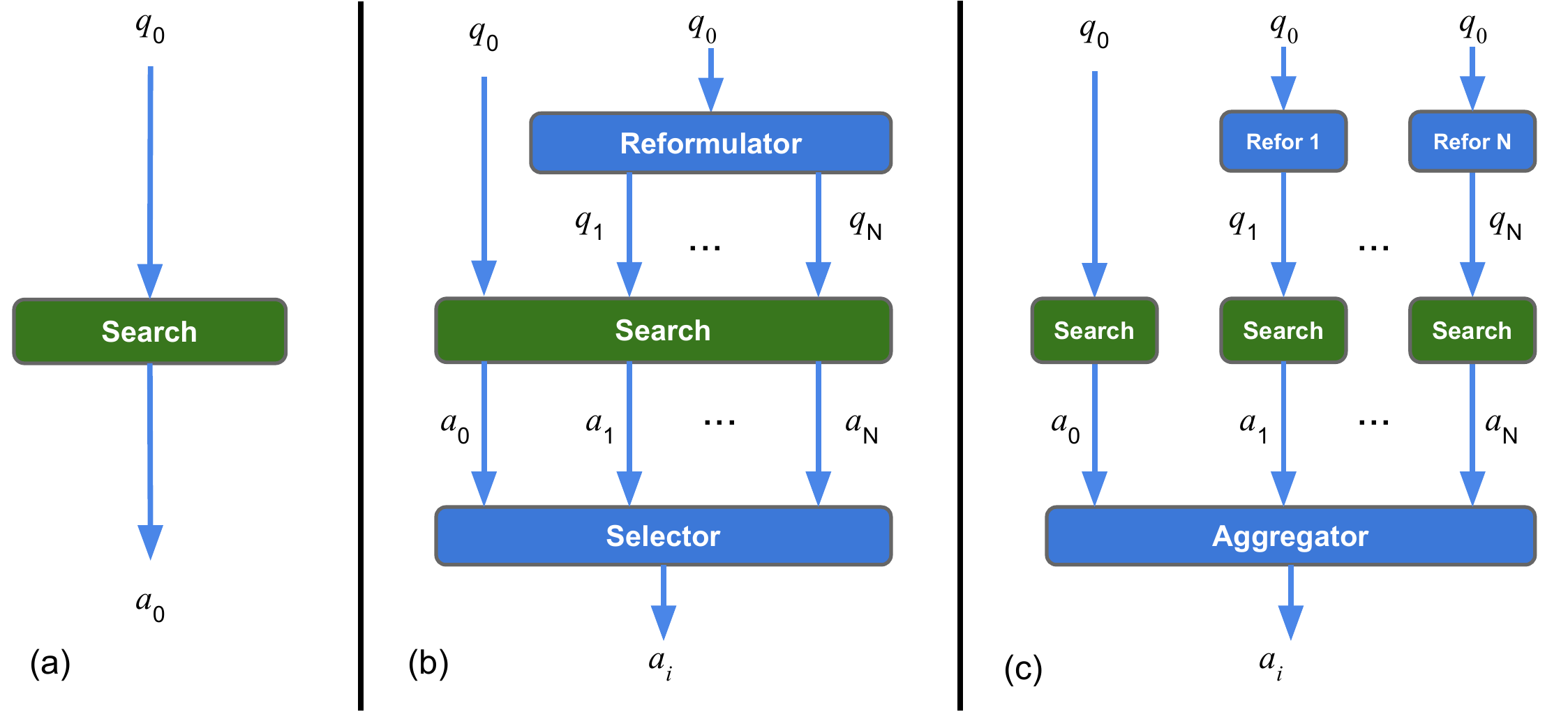}}
\vspace{-4mm}
\caption{\textbf{a)} A vanilla search system. The
  query $q_0$ is given to the system which outputs a result
  $a_0$. \textbf{b)} The search system with a reformulator. The
  reformulator queries the system with $q_0$ and its
  reformulations $\{q_1, ... q_N\}$ and receives back the results
  $\{a_0, ..., a_N\}$. A selector then decides the best result $a_i$ for
  $q_0$. \textbf{c)} The proposed system. The original query is
  reformulated multiple 
  times by different reformulators. Reformulations are used to
  obtain results from the search system, which are then sent to the
  aggregator, which picks the best result for the original query
  based on a learned weighted majority voting scheme. Reformulators
  are independently trained on disjoint partitions of the 
  dataset thus increasing the variability of reformulations.} 
\label{fig:overview}
\end{center}
\vspace{-6mm}
\end{figure*}

\subsection{Task}
We describe the proposed method using a generic end-to-end
search task. The problem consists in learning to reformulate a
query so that the underlying retrieval system can return a 
better result.

Following~\citet{nogueira2017task} and \citet{aqa-iclr:2018} we frame
the task as a reinforcement learning problem, in which the 
query reformulation system is an RL-agent that interacts with an
environment that provides answers and
rewards. The goal of the agent
is to generate reformulations such that the expected returned reward
(i.e., correct answers) is maximized. The environment is treated as
a black-box, i.e., the agent does not have direct access to any of its
internal mechanisms. 
Figure~\ref{fig:overview}-(b) illustrates this framework.

\subsection{System}
Figure~\ref{fig:overview}-(c) illustrates the new agent. An input
query $q_0$ is given to the $N$ sub-agents. A sub-agent is any system that
accepts as input a query and returns a corresponding
reformulation. Thus, sub-agents can be 
heterogeneous. 

Here we train each sub-agent on a partition of the training set. The $i$-th
agent queries the underlying search system with the reformulation $q_i$
and receives a result $a_i$. The set $\{(q_i, a_i) | 0 \leq i \leq N\}$ is
given to the aggregator, which then decides which result will be final.


\subsection{Sub-agents}
The first step for training the new agent is to partition the training
set. We randomly split it into equal-sized subsets. For an analysis of
how other partitioning methods affect performance, see
Appendix~\ref{sec:partitioning_methods}. 
In our implementation, a sub-agent is a sequence-to-sequence
model~\citep{sutskever2014sequence,cho2014learning} trained on a
partition of the dataset. It receives as an input the original
query $q_0$ and outputs a list of reformulated queries $(q_i)$
using beam search. 

Each reformulation $q_i$ is given to the same environment that returns
a list of results $(a_i^1, .., a_i^K)$ and their respective rewards
$(r_i^1,.. r_i^K)$. We then use REINFORCE~\citep{williams1992simple} to
train the sub-agent. At training time, instead of using beam search,
we sample reformulations.  

Note that we also add the identity agent (i.e., the reformulation is
the original query) to the pool of sub-agents. 

\subsection{Meta-Agent: Aggregator}
\label{sec:aggregator}
The aggregator receives as inputs $q_0$ and a list of candidate results
$(a_i^1,.. a_i^K)$ for each reformulation $q_i$. We first compute the
set of unique results ${a_j}$ and two different scores for each
result: the accumulated rank score $s_j^A$ and the relevance score $s_j^R$.  

The accumulated rank score is computed as $s_j^A = \sum_{i=1}^N
\frac{1}{\text{rank}_{i,j}}$, where $\text{rank}_{i,j}$ is the rank of the j-th
result when retrieved using $q_i$. The relevance score $s_j^R$ is the prediction that the result $a_j$ is
relevant to query $q_0$. It is computed as: 
\begin{equation}
s_j^R = \sigma(W_2 \text{ReLU}(W_1 z_j + b_1) + b_2),
\end{equation}
where
\begin{equation}
z_j = [f_{\text{CNN}}(q_0); f_{\text{BOW}}(a_j); f_{\text{CNN}}(q_0) - f_{\text{BOW}}(a_j); f_{\text{CNN}}(q_0) \odot f_{\text{BOW}}(a_j)], 
\label{eq:zi}
\end{equation}
$W_1 \in \RR^{4D \times D}$ and $W_2 \in \RR^{D \times 1}$ are weight
matrices, $b_1 \in \RR^{D}$ and $b_2 \in \RR^{1}$ are biases. The brackets in $[x; y]$ represent the concatenation of vectors $x$ and $y$. The symbol $\odot$ denotes the element-wise multiplication, $\sigma$ is the sigmoid function, and ReLU is a Rectified Linear
Unit function~\citep{nair2010rectified}. The function $f_{\text{CNN}}$ is
implemented as a CNN encoder\footnote{In the preliminary experiments, we found CNNs to work better than LSTMs~\citep{hochreiter1997long}.}
followed by average pooling over the
sequence~\citep{kim2014convolutional}. The function $f_{\text{BOW}}$ is the average word
embeddings of the result. At test time, the top-K answers with respect to $s_j={s_j^A s_j^R}$ are returned. 

We train the aggregator with stochastic gradient descent (SGD) to
minimize the cross-entropy loss: 
\begin{equation} \label{eq:aggregator_loss}
L = -\sum_{j \in J^*} \log (s_j^R) - \sum_{j \notin J^*} \log (1 - s_j^R),
\end{equation}
where $J^*$ is the set of indexes of the ground-truth results.
The architecture details and hyperparameters can be found in
Appendix~\ref{sec:hyperparameters}.

\section{Document Retrieval}

We now present experiments and results in a document retrieval task. In this task, the goal is to rewrite a query so that the number of relevant documents retrieved by a search engine increases. 

\subsection{Environment}

The environment receives a query as an action, and it returns a list of documents as an observation/state and a reward computed using a list of ground truth documents. We use Lucene\footnote{https://lucene.apache.org/} in its default configuration as our search engine, with BM25 as the ranking function. The input is a query and the output is a ranked list of documents. 

\subsection{Datasets}

To train and evaluate the models, we use three datasets:

\subparagraph{TREC-CAR:} Introduced by~\citet{dietz2017trec}, in this dataset the input query is the concatenation of a Wikipedia article title with the title of one of its section. The ground-truth documents are the paragraphs within that section. The corpus consists of all of the English Wikipedia paragraphs, except the abstracts. The released dataset has five predefined folds, and we use
the first four as a training set (approx. 3M queries), and the remaining as a validation set (approx. 700k queries). The test set is the same used evaluate the submissions to TREC-CAR 2017 (approx. 1,800 queries).

\subparagraph{Jeopardy:} This dataset was introduced
by~\citet{nogueira2016end}. The input is a \textit{Jeopardy!} question. The ground-truth document is a Wikipedia article whose title is the answer to the question. The corpus consists of all English Wikipedia articles. 

\subparagraph{MSA:} Introduced by~\citet{nogueira2017task}, this dataset consists of academic papers crawled from Microsoft Academic
API.\footnote{https://www.microsoft.com/cognitive-services/en-us/academic-knowledge-api} A query is the title of a paper and the ground-truth answer consists of the papers cited within. Each document in the corpus consists of its title and abstract. 

\subsection{Reward}
Since the main goal of query reformulation is to
increase the proportion of relevant documents returned, we use recall as the reward: $\text{R}@K
= \frac{|D_K \cap D^*|}{|D^*|}$, 
where $D_K$ are the top-$K$ retrieved documents and $D^*$ are the relevant documents. We also experimented using as a reward other metrics such as NDCG, MAP, MRR, and R-Precision but these resulted in similar or slightly worse performance than Recall@40. 
Despite the agents optimizing for Recall, we report the main results in MAP as this is a more commonly used metric in information retrieval. For results in other metrics, see Appendix~\ref{sec:other_metrics}.

\begin{table*}
\begin{center}
\begin{tabular}{l|ccc|cc}
 & \multirow{2}{*}{TREC-CAR} & \multirow{2}{*}{Jeopardy} & \multirow{2}{*}{MSA} & Training & FLOPs\\
 & & & & (Days) & ($\times 10^{18}$)\\
\noalign{\vskip 1mm}
\hline
\noalign{\vskip 1mm}
BM25 & 11.3 & 8.2 & 3.1 & \multicolumn{2}{c}{N/A} \\
PRF & 11.6 & 13.1 & 3.4 & \multicolumn{2}{c}{N/A} \\
RM3 & 12.0 & 13.5 & 3.1 & \multicolumn{2}{c}{N/A} \\
\noalign{\vskip 1mm}
\hline
\noalign{\vskip 1mm}
RL-RNN \citep{nogueira2017task} & 12.8 & 15.9 & 4.1 & 10 & 2.3\\
RL-10-Ensemble & 13.0 & 17.0 & 4.4 & 10 & 23.0\\
\noalign{\vskip 1mm}
\hline
\noalign{\vskip 1mm}
RL-10-Full & 14.1 & 29.3 & 4.9 & 1 & 2.3\\
RL-10-Bagging & 14.1 & 29.6 & 5.0 & 1 & 2.3\\
RL-10-Sub & 14.3 & 30.5 & 5.5 & 1 & 2.3\\
RL-10-Sub (Pretrained) & 14.4 & 30.7 & 5.4 & 10$^{\star}$+1 & 4.6\\
RL-10-Full (Extra Budget) & 14.8 & 31.2 & 5.6 & 10 & 23.0\\
RL-10-Full (Ensemble 10 Aggregators) & 17.7 & 33.9 & 6.1 & 10 & 23.0\\
\noalign{\vskip 1mm}
\hline
\noalign{\vskip 1mm}
RM3 + BERT Aggregator & 35.5 & 41.3 & 6.6 & 10 & 23.0\\
RL-10-Sub + BERT Aggregator & 36.4 & 42.5 & 7.2 & 10 & 23.0\\
\noalign{\vskip 1mm}
\hline
\noalign{\vskip 1mm}
Best System of TREC-CAR 2017 & 14.8 & - & - & - & -\\
\citep{macavaneycontextualized}
\end{tabular}
\end{center}
\vskip -1mm
\caption{MAP scores on the test sets of the document retrieval datasets. Similar results hold for other metrics (see Appendix~\ref{sec:other_metrics}). $^{\star}$The weights of the agents are initialized from a single model pretrained for ten days on the full training set.}
\label{tab:results_document_retrieval}
\end{table*}

\subsection{Baselines}

\subparagraph{BM25:} We give the original query to Lucene with BM25 as a ranking function and use the retrieved documents as results.

\subparagraph{PRF:} This is the pseudo relevance feedback
method~\citep{rocchio1971relevance}. We expand the original query with
terms from the documents retrieved by the Lucene search engine using the
original query. The top-N TF-IDF terms from each of the top-K
retrieved documents are added to the original query, where N and K are
selected by a grid search on the validation data. 

\subparagraph{Relevance Model (RM3):} This is our implementation of the relevance model for query expansion~\citep{lavrenko2001relevance}. The probability of adding a term $t$ to the original query is given by:
\begin{equation}
P(t|q_0) = (1-\lambda) P'(t|q_0) + \lambda \sum_{d \in D_0} P(d) P(t|d) P(q_0|d),
\end{equation}
where $P(d)$ is the probability of retrieving the document $d$, assumed uniform over the set, $P(t|d)$ and $P(q_0|d)$ are the probabilities assigned by the language model obtained from $d$ to $t$ and $q_0$, respectively. $P'(t|q_0)= \frac{\text{tf}(t \in q)}{|q|}$, where $\text{tf}(t,d)$ is the term frequency of $t$ in $d$. We set the interpolation parameter $\lambda$ to 0.65, which was the best value found by a grid-search on the development set.

We use a Dirichlet smoothed language model~\citep{zhai2001study} to compute a language model from a document $d \in D_0$:
\begin{equation}
P(t|d)=\frac{\text{tf}(t,d)+u P(t|C)}{|d| + u},
\end{equation}
where $u$ is a scalar constant ($u=1500$ in our experiments), and $P(t|C)$ is the probability of $t$ occurring in the entire corpus $C$.

We use the $N$ terms with the highest $P(t|q_0)$ in an expanded query, where $N=100$ was the best value found by a grid-search on the development set. 

\subparagraph{RL-RNN:} This is the sequence-to-sequence model trained with reinforcement learning from~\citet{nogueira2017task}. The reformulated query is formed by appending new terms to the original query. The terms are selected from the documents retrieved using the original query. The agent is trained from scratch. 

\subparagraph{RL-N-Ensemble:} We train $N$ RL-RNN agents with different initial weights on the full training set. At test time, we average the probability distributions of all the $N$ agents at each time step and select the token with the highest probability, as done by~\citet{sutskever2014sequence}.

\subsection{Proposed Models}

We evaluate the following variants of the proposed method: 
\subparagraph{RL-N-Full:} We train $N$ RL-RNN agents with different
initial weights on the full training set. The answers are obtained using the best (greedy) reformulations of all the agents and are given to the aggregator.

\subparagraph{RL-N-Bagging:} This is the same as RL-N-Full but we construct the training set of each RL-RNN agent by sampling with replacement D times from the full training set, which has a size of D. This is known as the bootstrap sample and leads to approximately 63\% unique samples, the rest being duplicates.

\subparagraph{RL-N-Sub:} This is the proposed agent. It is similar to RL-N-Full but the multiple sub-agents are trained on random partitions of the
dataset (see Figure~\ref{fig:overview}-(c)).

\subparagraph{BERT Aggregator:} We experimented replacing our simple aggregator with BERT~\citep{devlin2018bert}, which recently achieved the state-of-the-art in a wide range of textual tasks. Using the same notation used in their paper, we feed the query as sentence A and the document text as sentence B. We truncate the document text such that concatenation of query, document, and separator tokens have a maximum length of 512 tokens. We use a pretrained $\text{BERT}_\text{LARGE}$ model as a binary classification model, that is, we feed the $[\text{CLS}]$ vector to a single layer neural network and obtain the probability of the document being correct. We obtain the final list of documents by ranking them with respect these probabilities. We train it with the same objective used to train our aggregator (Equation \ref{eq:aggregator_loss}). 
To compare how well our proposed reformulation agents perform against the best non-neural reformulation method, we implemented two variants of the system described here. One is when the initial list of candidate documents $a_j$ is given by RM3 (RM3 + BERT Aggregator), and the other is by RL-10-Sub (RL-10-Sub + BERT Aggregator).

\subsection{Results}
A summary of the document retrieval results is shown in
Table~\ref{tab:results_document_retrieval}. We estimate the number of floating point operations used to train a model by multiplying the training time, the number of GPUs used, and 2.7 TFLOPS as an estimate of the single-precision floating-point of a K80 GPU.

Since the sub-agents are frozen during the training of the aggregator, we pre-compute all $(q_0, q_i, a_i, r_i)$ tuples from the training set, thus avoiding sub-agent or environment calls. This reduces its training time to less than 6 hours ($0.06\times10^{18}$ FLOPs). Since this cost is negligible when compared to the sub-agents', we do not include it in the table.

The proposed methods (RL-10-\{Sub, Bagging, Full\}) have 20-60\% relative
performance improvement over the standard ensemble (RL-10-Ensemble)
while training ten times faster. More interestingly, RL-10-Sub has a better performance than the single-agent version (RL-RNN), uses the same computational budget, and trains on a fraction of the time. Lastly, we found that RL-10-Sub (Pretrained) has the best balance between performance and training cost across all datasets.

Compared to the top-performing system in the TREC-CAR 2017 Track~\citep{macavaneycontextualized}, an RL-10-Full with an ensemble of 10 aggregators yields a relative performance improvement of approximately 20\%.

By replacing our aggregator with BERT, we improve performance by 50-100\% in all three datasets (RL-10-Sub + BERT Aggregator). This is a remarkable improvement given that we used BERT without any modification from its original implementation. Without using our reformulation agents, the performance drops by 3-10\% (RM3 + BERT Aggregator).

For an analysis of the aggregator's contribution to the overall performance, see Appendix~\ref{sec:analysis_aggregator}.

\begin{figure*}
\begin{center}
\centerline{
  \includegraphics[width=0.32\textwidth]{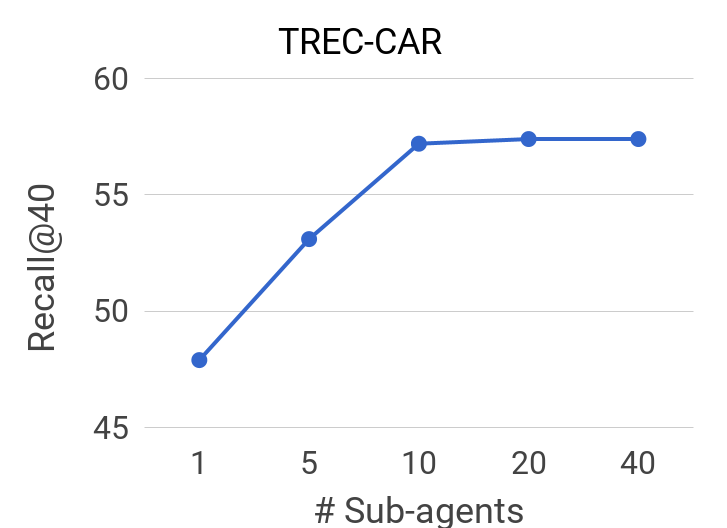}
  \includegraphics[width=0.3\textwidth]{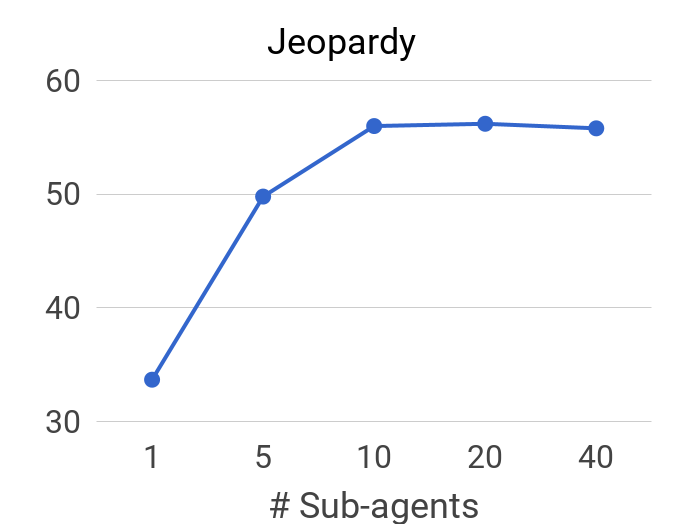}
  \includegraphics[width=0.3\textwidth]{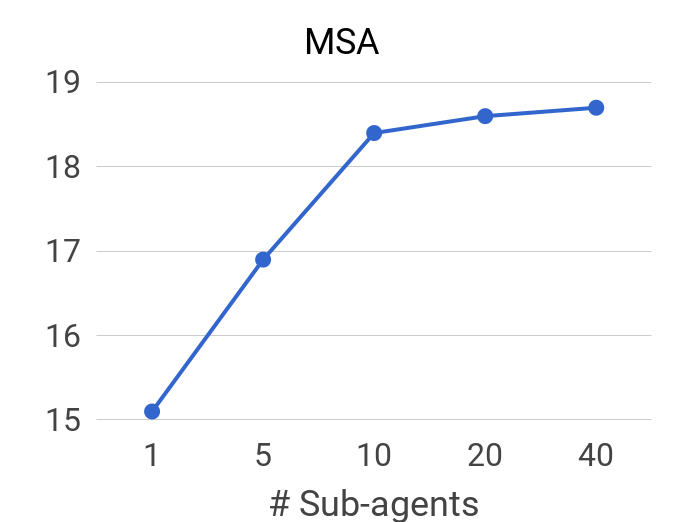}
}
\vskip -1mm
\caption{Overall system's performance for different number of sub-agents.}
\label{fig:number_sub_agents}
\end{center}
\vskip -8mm
\end{figure*}

\subparagraph{Number of Sub-Agents:} We compare the performance of the full system (reformulators + aggregator) for different numbers of agents in Figure~\ref{fig:number_sub_agents}. The performance of the system is stable across all datasets after more than ten sub-agents are used, thus indicating the robustness of the proposed method. For more experiments regarding training stability, see Appendix~\ref{sec:training_stability}.

\section{Question-Answering}

To further assess the effectiveness of the proposed method, we conduct
experiments in a question-answering task, comparing our agent with
the active question answering agent proposed by~\citet{aqa-iclr:2018}.


The environment receives a question as an action and returns an
answer as an observation, and a reward computed against a
ground truth answer. We use either BiDAF~\citep{seo2016bidirectional} or BERT~\citep{devlin2018bert} as a question-answering system. Given a
question, it outputs an answer span from a 
list of snippets. We use as a reward the token level F1 score on the
answer (see Section~\ref{sec:evaluation_metrics} for its definition).  

We follow~\citet{aqa-iclr:2018} to train BiDAF and BERT. We
emphasize that their parameters are frozen when we train and evaluate
the reformulation system. Training and evaluation are performed on the SearchQA
dataset~\citep{dunn2017searchqa}. The data contains \textit{Jeopardy!} clues
as questions. Each clue has a correct answer and a list of 50 snippets
from Google's top search results. The training, validation and test
sets contain 99,820, 13,393 and 27,248 examples, respectively.

\subsection{Baselines and Benchmarks}
We compare our agent against the following baselines and benchmarks:

\subparagraph{BiDAF/BERT:} The original question is given to the
question-answering system without any modification (see
Figure~\ref{fig:overview}-(a)).


\subparagraph{AQA:} This is the best model from~\citet{aqa-iclr:2018}. It
consists of a reformulator and a selector. The reformulator is a
subword-based sequence-to-sequence model that produces twenty
reformulations of an input question using beam search. The answers for the original question and its reformulations are obtained from BiDAF. These are given to the selector which then chooses one of the answers as final (see
Figure~\ref{fig:overview}-(b)). The reformulator is pretrained on
translation and paraphrase data. 

\begin{table*}
\begin{center}
\begin{tabular}{l|cc|cc|cc}
 & \multicolumn{2}{c}{Dev} & \multicolumn{2}{|c|}{Test} & Training & FLOPs\\
 & F1 & Oracle & F1 & Oracle & (Days) & ($\times 10^{18}$) \\
\noalign{\vskip 1mm}
\hline
\noalign{\vskip 1mm}
BiDAF~\citep{seo2016bidirectional} & 37.9 & - & 34.6 & - & \multicolumn{2}{c}{N/A} \\
R$^3$~\citep{wang2017r} & - & - & 55.3 & - & \multicolumn{2}{c}{N/A} \\
Re-Ranker~\citep{wang2017evidence} & - & - & 60.6 & - & \multicolumn{2}{c}{N/A}\\
DS-QA~\citep{lin2018denoising} & - & - & 64.5 & - & \multicolumn{2}{c}{N/A}\\
BERT (our run) & 71.2 & - & 69.1 & - & \multicolumn{2}{c}{N/A} \\
\noalign{\vskip 1mm}
\hline
\noalign{\vskip 1mm}
AQA~\citep{aqa-iclr:2018} & 47.4 & 56.0 & 45.6 & 53.8 & 10 & 4.6 \\
\noalign{\vskip 1mm}
\hline
\noalign{\vskip 1mm}
BiDAF + AQA-10-Sub & 51.7 & 66.8 & 49.0 & 61.5 & 1 & 4.6 \\
BiDAF + AQA-10-Full & 51.0 & 61.2 & 48.4 & 58.7 & 1 & 4.6 \\
BiDAF + AQA-10-Full (extra budget) & 51.4 & 61.3 & 50.5 & 58.9 & 10 & 46.0 \\
\noalign{\vskip 1mm}
\hline
\noalign{\vskip 1mm}
BERT + AQA-10-Sub & 71.2 & 76.8 & 69.1 & 75.4 & 1 & 4.6 \\
\end{tabular}
\end{center}
\vskip -1mm
\caption{Main result on the question-answering task (SearchQA dataset). We did not include the training cost of the aggregator (0.2 days, 0.06 $\times 10^{18}$ FLOPs).
}
\label{tab:results_question_answering}
\vspace{-4mm}
\end{table*}

\subsection{Proposed Methods}

\subparagraph{AQA-N-\{Full, Sub\}:} Similar to the RL-N-\{Full, Sub\} models, we use AQA reformulators as the sub-agents followed by an aggregator to create AQA-N-Full and AQA-N-Sub models, whose sub-agents are trained on the full and random partitions of the dataset, respectively. For the training and hyperparameter details, see Appendix~\ref{sec:hyperparameters_question_answering}.

\subsection{Evaluation Metrics}
\label{sec:evaluation_metrics}

\subparagraph{F1:} We use the macro-averaged F1 score as the main metric. It measures the average bag of tokens overlap between the prediction and ground truth answer. We take the F1 over the ground truth answer for a given question and then average over all of the questions.

\subparagraph{Oracle:} Additionally, we present the oracle performances, which are from a perfect aggregator that predicts $s_j^R=1$ for relevant answers and $s_j^R=0$, otherwise.

\subsection{Results}
\label{sec:question_answering_results}

Results are presented in Table~\ref{tab:results_question_answering}. When using BiDAF as the underlying Q\&A system, the proposed method (AQA-10-\{Full, Sub\}) have both better F1 and oracle performances than the single-agent AQA method, while training in one-tenth of the time. Even when the ensemble method is given ten times more training time (AQA-10-Full, extra budget), our method achieves a higher performance.

We achieve the state-of-the-art on SearchQA by using BERT without any modification from its original implementation. Our reformulation strategy (BERT + AQA-10-Sub), however, could not improve upon this underlying Q\&A system. We conjecture that, although there is room for improvement, as the oracle performance is 5-7\% higher than BERT alone, the reformulations and answers do not contain enough information for the aggregator to discriminate good from bad answers. One possible way to fix this is to give the context of the answer to the aggregator, although in our experiments we could not find any successful way to use this extra information. 

\subparagraph{Original Query Contribution:} We observe a drop in F1 of approximately 1\% when the original query is removed from the pool of reformulations, which shows that the gains come mostly from the multiple reformulations and not from the aggregator falling back on selecting the original query.

\subsection{Query Diversity} 
\label{sec:query_diversity}

Here we evaluate how query diversity and performance are related. For that, we use four metrics (defined in Appendix~\ref{sec:query_diversity_metrics}): pCos, pBLEU, PINC, and Length Std.

Table~\ref{tab:diversity} shows that the multiple agents trained on partitions of the dataset (AQA-10-Sub) produce more diverse queries than a single agent with beam search (AQA) and multiple agents trained on the full training set (AQA-10-Full). This suggests that its higher performance can be partly attributed to the higher diversity of the learned policies.

\begin{table*}
\begin{center}
\begin{tabular}{l|cccc|cc}
Method & pCos $\downarrow$ & pBLEU $\downarrow$ & PINC $\uparrow$ & Length Std $\uparrow$ & F1 $\uparrow$ & Oracle $\uparrow$\\
\noalign{\vskip 1mm}
\hline
\noalign{\vskip 1mm}
AQA & 66.4 & 45.7 & 58.7 & 3.8 & 47.7 & 56.0\\
AQA-10-Full & 29.5 & 26.6 & 79.5 & 9.2 & 51.0 & 61.2\\
AQA-10-Sub & \textbf{14.2} & \textbf{12.8} & \textbf{94.5} & \textbf{11.7} & \textbf{51.4} & \textbf{61.3} \\
\end{tabular}
\end{center}
\vspace{-2mm}
\caption{Diversity scores of reformulations from different methods. For pBLEU and pCos, lower values mean higher diversity. Notice that higher diversity scores are associated with higher F1 and oracle scores.}
\label{tab:diversity}
\end{table*}

\section{Conclusion}

We proposed a method to build a better query reformulation system by training multiple sub-agents on partitions of the data using reinforcement learning and an aggregator that learns to combine the answers of the multiple agents given a new query. We showed the effectiveness and efficiency of the proposed approach on the tasks of document retrieval and question answering. One interesting orthogonal extension would be to introduce diversity on the beam search decoder~\citep{vijayakumar2016diverse,li2016simple}, thus shedding light on the question of whether the gains come from the increased capacity of the system due to the use of the multiple agents, the diversity of reformulations, or both. 

\bibliographystyle{iclr2019_conference}
\bibliography{main}

\newpage
\begin{appendices}

\section{Document Retrieval: Results On More Metrics}
\label{sec:other_metrics}
Following~\cite{dietz2017trec}, we report the results on four standard TREC evaluation measures: R-Precision (R-Prec), Mean-average Precision (MAP), Reciprocal Rank (MRR), and Normalize Discounted Cumulative Gain (NDCG). We also include Recall@40 as this is the reward our agents are optimizing for. The results for TREC-CAR, Jeopardy, and MSA are in Tables~\ref{tab:results_trec_car}, \ref{tab:results_jeopardy}, \ref{tab:results_msa}, respectively.

\section{Hyperparameters}
\label{sec:hyperparameters}

\subsection{Document Retrieval Task}

\subparagraph{Sub-agents:} We use mini-batches of size 256, ADAM~\citep{kingma2014adam} as the optimizer, and learning rate of $10^{-4}$.
\subparagraph{Aggregator:} The encoder $f_{q_0}$ is a word-level two-layer CNN with filter sizes of 9 and 3, respectively, and 128 and 256 kernels, respectively. $D=512$. No dropout is used. ADAM is the optimizer with learning rate of $10^{-4}$ and mini-batch of size 64. It is trained for 100 epochs.

\subsection{Question-Answering Task}
\label{sec:hyperparameters_question_answering}
\subparagraph{Sub-agents:} We use mini-batches of size 64, SGD as the optimizer, and learning rate of $10^{-3}$.
\subparagraph{Aggregator:} The encoder $f_{q_0}$ is a token-level, three-layer CNN with filter sizes of 3, and 128, 256, and 256 kernels, respectively. We train it for 100 epochs with mini-batches of size 64 with SGD and learning rate of $10^{-3}$.

\section{Aggregator Analysis}
\label{sec:analysis_aggregator}

\subsection{Contribution of the Aggregator vs. Multiple Reformulators}

To isolate the contribution of the Aggregator from the gains brought by the multiple reformulators, we use the aggregator to re-rank the list of documents obtained with the rewrite from a single reformulator (RL-RNN Greedy + Aggregator). We also use beam search or sampling to produce $K$ rewrites from a single reformulator (RL-RNN $K$ Sampled/Beam + Aggregator). The $K$ lists of ranked documents returned by the environment are then merged into a single list and re-ranked by the Aggregator.

The results are shown in table~\ref{tab:aggregators_contribution}. The higher performance obtained with ten rewrites produced by different reformulators (RL-10-Sub) when compared 20 sampled rewrites from a single agent (RL-RNN 20 Sampled + Aggregator) indicates that the gains the proposed method comes mostly from the pool of diverse reformulators, and not from the simple use of a re-ranking function (Aggregator).

\subsection{Ablation Study}
To validate the effectiveness of the proposed aggregation function, we conducted a comparison study on the TREC-CAR dataset. We present the results in Table~\ref{tab:comparison}. We notice that removing or changing the accumulated rank or relevance score functions results in a performance drop between 0.4-1.4\% in MAP. The largest drop occurs when we remove the aggregated rank ($s_j=s^R_j$), suggesting that the rank of a document obtained from the reformulation phase is a helpful signal to the re-ranking phase.

Not reported in the table, we also experimented concatenating to the input vector $z_i$ (eq.~\ref{eq:zi}) a vector to represent each sub-agent. These vectors were learned during training and allowed the aggregator to distinguish sub-agents. However, we did not notice any performance improvement.

\begin{table}
\begin{center}
\begin{tabular}{l|ccccc}
 & R@40 & MAP & R-Prec & MRR & NDCG\\
\noalign{\vskip 1mm}
\hline
\noalign{\vskip 1mm}
BM25 & 27.5 & 11.3 & 9.8 & 21.0 & 27.4\\
PRF & 28.6 & 11.7 & 10.1 & 21.7 & 27.2\\
RM3 & 29.7 & 12.1 & 10.5 & 22.5 & 27.2\\
\noalign{\vskip 1mm}
\hline
\noalign{\vskip 1mm}
RL-RNN & 31.6 & 12.7 & 10.9 & 23.6 & 28.3\\
RL-10-Ensemble & 31.7 & 12.8 & 11.0 & 23.8 & 28.3\\
\noalign{\vskip 1mm}
\hline
\noalign{\vskip 1mm}
RL-RNN Greedy + Aggregator & 32.0 & 12.8 & 11.1 & 23.2 & 29.0 \\
RL-RNN 20 Sampled + Aggregator & 32.5 & 12.0 & 11.2 & 23.1 & 29.3\\
RL-RNN 20 Beam + Aggregator & 32.3 & 12.9 & 11.1 & 23.0 & 29.2\\
\noalign{\vskip 1mm}
\hline
\noalign{\vskip 1mm}
RL-10-Full & 35.2 & 14.1 & 12.0 & 24.5 & 29.5\\
RL-10-Bagging & 35.9 & 14.1 & 12.1 & 24.6 & 29.7\\
RL-10-Sub & 36.7 & 14.2 & 12.1 & 25.0 & 29.5\\
RL-10-Sub (Pretrained) & 36.9 & 14.4 & 12.3 & 25.0 & 29.8\\
RL-10-Full (Extra Budget) & 37.7 & 14.8 & 12.5 & 25.3 & 29.1\\
RL-10-Full (Ensemble of 10 Aggregators) & 39.5 & 17.7 & 13.5 & 26.3 & 30.8\\
\noalign{\vskip 1mm}
\hline
\noalign{\vskip 1mm}
RM3 + BERT Aggregator & 50.9 & 35.5 & 32.1 & 51.1 & 45.2\\
RL-10-Full + BERT Aggregator & 52.0 & 36.4 & 32.6 & 52.0 & 46.6\\
\noalign{\vskip 1mm}
\hline
\noalign{\vskip 1mm}
Best System of TREC-CAR 2017 & - & 14.8 & 11.6 & 22.3 & 22.8\\
\citep{macavaneycontextualized} \\
\end{tabular}
\end{center}
\vskip -1mm
\caption{Results on more metrics on the test set of the TREC-CAR dataset.}
\label{tab:results_trec_car}
\end{table}

\begin{table}
\begin{center}
\begin{tabular}{l|ccccc}
 & R@40 & MAP & R-Prec & MRR & NDCG\\
\noalign{\vskip 1mm}
\hline
\noalign{\vskip 1mm}
BM25 & 23.0 & 8.2 & 4.4 & 8.2 & 11.9 \\
PRF & 29.7 & 13.1 & 8.4 & 13.1 & 17.4\\
RM3 & 30.5 & 13.5 & 8.7 & 13.5 & 17.9\\
\noalign{\vskip 1mm}
\hline
\noalign{\vskip 1mm}
RL-RNN & 33.7 & 15.9 & 10.6 & 15.9 & 20.5\\
RL-10-Ensemble & 35.2 & 17.0 & 11.4 & 17.0 & 21.8\\
\noalign{\vskip 1mm}
\hline
\noalign{\vskip 1mm}
RL-RNN Greedy + Aggregator & 42.0 & 22.1 & 15.5 & 22.1 & 27.2 \\
RL-RNN 20 Sampled + Aggregator & 42.4 & 22.4 & 15.7 & 22.4 & 27.5\\
RL-RNN 20 Beam + Aggregator & 42.3 & 22.3 & 15.6 & 22.3 & 27.3 \\
\noalign{\vskip 1mm}
\hline
\noalign{\vskip 1mm}
RL-10-Full & 52.1 & 29.3 & 21.1 & 29.3 & 35.1\\
RL-10-Bagging & 52.5 & 29.6 & 21.4 & 29.6 & 35.4\\
RL-10-Sub & 53.5 & 29.7 & 23.0 & 30.6 & 36.4\\
RL-10-Sub (Pretrained) & 54.0 & 30.7 & 22.2 & 30.7 & 36.6 \\
RL-10-Full (Extra Budget) & 54.4 & 31.2 & 22.7 & 31.2 & 37.2\\
\noalign{\vskip 1mm}
\hline
\noalign{\vskip 1mm}
RM3 + BERT Aggregator & 61.5 & 41.3 & 36.1 & 41.3 & 43.4\\
RL-10-Full + BERT Aggregator & 62.7 & 42.5 & 37.0 & 42.5 & 44.8\\
\end{tabular}
\end{center}
\vskip -1mm
\caption{Results on more metrics on the test set of the Jeopardy dataset.}
\label{tab:results_jeopardy}
\end{table}

\begin{table}
\begin{center}
\begin{tabular}{l|ccccc}
 & R@40 & MAP & R-Prec & MRR & NDCG\\
\noalign{\vskip 1mm}
\hline
\noalign{\vskip 1mm}
BM25 & 12.7 & 3.1 & 6.0 & 15.4 & 9.1 \\
PRF & 13.2 & 3.4 & 6.4 & 16.2 & 9.7\\
RM3 & 12.3 & 3.1 & 6.0 & 15.0 & 8.9\\
\noalign{\vskip 1mm}
\hline
\noalign{\vskip 1mm}
RL-RNN & 15.1 & 4.1 & 7.3 & 18.8 & 11.2\\
RL-10-Ensemble & 15.8 & 4.4 & 7.7 & 19.7 & 11.7\\
\noalign{\vskip 1mm}
\hline
\noalign{\vskip 1mm}
RL-RNN Greedy + Aggregator & 16.1 & 4.5 & 7.8 & 20.1 & 12.0\\
RL-RNN 20 Sampled + Aggregator & 16.4 & 4.6 & 7.9 & 20.5 & 12.2\\
RL-RNN 20 Beam + Aggregator & 16.2 & 4.5 & 7.9 & 20.3 & 12.1\\
\noalign{\vskip 1mm}
\hline
\noalign{\vskip 1mm}
RL-10-Full & 17.4 & 4.9 & 8.4 & 21.9 & 13.0\\
RL-10-Bagging & 17.6 & 5.0 & 8.5 & 22.1 & 13.2\\
RL-10-Sub & 18.9 & 5.5 & 9.2 & 23.9 & 14.2\\
RL-10-Sub (Pretrained) & 19.1 & 5.4 & 9.1 & 24.0 & 14.2 \\
RL-10-Full (Extra Budget) & 19.2 & 5.6 & 9.3 & 24.3 & 14.4\\
\noalign{\vskip 1mm}
\hline
\noalign{\vskip 1mm}
RM3 + BERT Aggregator & 22.7 & 6.6 & 8.9 & 33.0 & 16.2\\
RL-10-Full + BERT Aggregator & 23.8 & 7.2 & 10.2 & 34.7 & 17.6\\
\end{tabular}
\end{center}
\vskip -1mm
\caption{Results on more metrics on the test set of the MSA dataset.}
\label{tab:results_msa}
\end{table}

\begin{table}
\begin{center}
\begin{tabular}{l|ccc}
 & TREC-CAR & Jeopardy & MSA\\
\noalign{\vskip 1mm}
\hline
\noalign{\vskip 1mm}
RL-RNN & 10.8 & 15.0 & 4.1\\
\noalign{\vskip 1mm}
\hline
\noalign{\vskip 1mm}
RL-RNN Greedy + Aggregator & 10.9 & 21.2 & 4.5\\
RL-RNN 20 Sampled + Aggregator & 11.1 & 21.5 & 4.6\\
RL-RNN 20 Beam + Aggregator & 11.0 & 21.4 & 4.5\\
\noalign{\vskip 1mm}
\hline
\noalign{\vskip 1mm}
RL-10-Sub & 12.3 & 29.7 & 5.5\\
\end{tabular}
\end{center}
\vskip -1mm
\caption{Multiple reformulators vs. aggregator contribution. Numbers are MAP scores on the dev set. Using a single reformulator with the aggregator (RL-RNN Greedy/Sampled/Beam + Aggregator) improves performance by a small margin over the single reformulator without the aggregator (RL-RNN). Using ten reformulators with the aggregator (RL-10-Sub) leads to better performance, thus indicating that the pool of diverse reformulators is responsible for most of the gains of the proposed method.}
\label{tab:aggregators_contribution}
\end{table}

\begin{table}
\begin{center}
\begin{tabular}{l|cc}
Aggregator Function & MAP & Diff \\
\noalign{\vskip 1mm}
\hline
$s_j = s_j^A s_j^R$ (proposed, Section~\ref{sec:aggregator}) & 12.3 & - \\
$z_j=f_{\text{CNN}}(q_0)||f_{\text{BOW}}(a_j)$ (eq.~\ref{eq:zi}) & 11.9 & -0.4 \\
$s_j^A = \sum_{i=1}^N \mathbbm{1}_{a_i=a_j}$ & 11.7 & -0.6 \\
$s_j = s_j^A$ & 11.1 & -1.2 \\
$s_j = s_j^R$ & 10.9 & -1.4 \\
\end{tabular}
\end{center}
\vskip -1mm
\caption{Comparison of different aggregator functions on TREC-CAR. The reformulators are from RL-10-Sub.}
\label{tab:comparison}
\vspace{-4mm}
\end{table}

\section{Training Stability of Single vs. Multi-Agent}
\label{sec:training_stability}
Reinforcement learning algorithms that use non-linear function approximators, such as neural networks, are known to be unstable \citep{tsitsiklis1996analysis, fairbank2011divergence, pirotta2013adaptive, mnih2015human}. Ensemble methods are known to reduce this variance~\citep{freund1995boosting,breiman1996bagging,breiman1996bias}. Since the proposed method can be viewed as an ensemble, we compare the AQA-10-Sub's F1 variance against a single agent (AQA) on ten runs. Our method has a much smaller variance: 0.20 vs. 1.07. We emphasize that it also has a higher performance than the AQA-10-Ensemble.

We argue that the higher stability is due to the use of multiple agents. Answers from agents that diverged during training can be discarded by the aggregator. In the single-agent case, answers come from only one, possibly bad, policy.

\section{Diversity Metrics}
\label{sec:query_diversity_metrics}

Here we define the metrics used in query diversity analysis (Sec.~\ref{sec:query_diversity}):

\subparagraph{pCos:} Mean pair-wise cosine distance: $\frac{1}{N} \sum_{n=1}^N \frac{1}{|Q^n|} \sum_{q, q' \in Q^n} \text{cos}\big(\#q, \#q'\big)$,
where $Q^n$ is a set of reformulated queries for the $n$-th original query in the development set and $\#q$ is the token count vector of q.

\subparagraph{pBLEU:} Mean pair-wise sentence-level BLEU \citep{chen2014systematic}: 
$\frac{1}{N} \sum_{n=1}^N \frac{1}{|Q^n|} \sum_{q, q' \in Q^n} \text{BLEU}\big(q, q'\big)$.

\subparagraph{PINC}: Mean pair-wise paraphrase in k-gram changes~\citep{chen2011collecting}: 
$\frac{1}{N} \sum_{n=1}^N \frac{1}{|Q^n|} \sum_{q, q' \in Q^n} \frac{1}{K} \sum_{k=1}^K 1 - \frac{|\text{k-gram}_q \cap \text{k-gram}_{q'}|}{|\text{k-gram}_{q'}|},$
where $K$ is the maximum number of k-grams considered (we use $K=4$).

\subparagraph{Length Std:} Standard deviation of the reformulation lengths: $\frac{1}{N} \sum_{n=1}^N \text{std}\big(\{|q_i^n|\}_{i=1}^{|Q|}\big)$

\section{On Data Partitioning}
\label{sec:partitioning_methods}

\begin{table}
\small
\begin{center}
\begin{tabu}{l|cccc|cccc}
& \multicolumn{4}{c|}{SearchQA} & \multicolumn{4}{c}{TREC-CAR} \\
\rowfont{\tiny}
& $E_i[e_i]\downarrow$ & $E_i[E_{j\neq i}[s_{ij}]]\uparrow$ & $E_i[V_{j\neq i}[s_{ij}]]\downarrow$ & F1$\uparrow$ & $E_i[e_i]\downarrow$ & $E_i[E_{j\neq i}[s_{ij}]]\uparrow$ & $E_i[V_{j\neq i}[s_{ij}]]\downarrow$ & R@40$\uparrow$ \\
\noalign{\vskip 1mm}
\hline
\noalign{\vskip 1mm}
Q & 9.9 & 52.0 & \textbf{1.1} & 53.3 & 15.3 & 50.4 & 5.9 & 50.0 \\
A & 22.0 & 50.1 & 3.9 & 51.4 & \textbf{1.3} & \textbf{57.0} & 0.3 & \textbf{56.9} \\
Q+A & \textbf{9.0} & 50.5 & 1.2 & \textbf{53.4} & 1.8 & 56.2 & 0.3 & 56.5\\
Rand. & 9.5 & \textbf{53.8} & \textbf{1.1} & \textbf{53.4} & 1.9 & \textbf{57.0} & \textbf{0.2} & \textbf{57.1}\\
\end{tabu}

\end{center}
\vskip -1mm
\caption{Partitioning strategies and the corresponding evaluation metrics. We notice that the random strategy generally results in the best quality sub-agents, leading to the best scores on both of the tasks.}
\label{tab:partitioning_results}
\vspace{-4mm}
\end{table}

Throughout this paper, we used sub-agents trained on random partitions of the dataset. We now investigate how different data partitioning strategies affect final performance of the system. Specifically, we compare the random split against a mini-batch K-means clustering algorithm~\citep{sculley2010web}.

\paragraph{Balanced K-means Clustering} 

For K-means, we experimented with three types of features: average question embedding (Q), average answer embedding (A), and the concatenation of these two (Q+A). The word embeddings were obtained from~\citet{mikolov2013efficient}.

The clusters returned by the K-means can be highly unbalanced. This is undesirable since some sub-agents might end up being trained with too few examples and thus may have a worse generalization performance than the others. To address this problem, we use a greedy cluster balancing algorithm as a post-processing step (see Algorithm~\ref{cluster_balancing} for the pseudocode).

\begin{algorithm}
\caption{Cluster Balancing}
\label{cluster_balancing}

\begin{algorithmic}[1] 

\State Given: desired cluster size $M$, and a set of clusters $C$, each containing a set of items.
\State sort $C$ by descending order of sizes
\State $C_\text{remaining} \gets \text{shallow\_copy}(C)$

\For{c in C}

  \State remove $c$ from $C_\text{remaining}$
  
  \While{$\text{c.size} < M$}
    \State $\text{item} \gets \text{randomly select an item from c}$
    \State move item to the closest cluster in $C_\text{remaining}$
    \State sort $C_\text{remaining}$ by descending order of sizes
  \EndWhile
\EndFor
\State \textbf{return} $C$
\end{algorithmic}
\end{algorithm}

\paragraph{Evaluation Metric}

In order to gain insight into the effect of a partitioning strategy, we first define three evaluation metrics. Let $\pi_i$ be the $i$-th sub-agent trained on the $i$-th partition out of $K$ partitions obtained from clustering. We further use $s_{ij}$ to denote the score, either F-1 in the case of question answering or R@40 for document retrieval, obtained by the $i$-th sub-agent $\pi_i$ on the $j$-th partition. 

{\bf Out-of-partition score} computes the generalization capability of the sub-agents outside the partitions on which they were trained:
\[
E_{i} [E_{j \neq i}[s_{ij}]] = \frac{1}{N} \sum_{i=1}^N \frac{1}{K-1} \sum_{j \neq i} s_{ij}.
\]
This score reflects the general quality of the sub-agents.
{\bf Out-of-partition variance} computes how much each sub-agent's performance on the partitions, on which it was not trained, varies:
\begin{equation}
\small
E_{i} [V_{j \neq i}[s_{ij}]] = \frac{1}{N} \sum_{i=1}^N \frac{1}{K-2} \sum_{j \neq i} (s_{ij} - E_{i \neq j}[s_{ij}])^2.
\end{equation}
It indicates the general stability of the sub-agents. If it is high, it means that the sub-agent must be carefully combined in order for the overall performance to be high.
{\bf Out-of-partition error} computes the generalization gap between the partition on which the sub-agent was trained and the other partitions:
\[
E_i[e_{i}] = \frac{1}{N} \sum_{i=1}^N (s_ij - E_{j \neq i}[s_{ij}]).
\]
This error must be low, and otherwise, would indicate that each sub-agent has overfit the particular partition, implying the worse generalization.


\paragraph{Result} 

We present the results in Table~\ref{tab:partitioning_results}. Although we could obtain a good result with the clustering-based strategy, we notice that this strategy is highly sensitive to the choice of features. Q+A is optimal for SearchQA, while A is for TREC-CAR. On the other hand, the random strategy performs stably across both of the tasks, making it a preferred strategy. Based on comparing Q and Q+A for SearchQA, we conjecture that it is important to have sub-agents that are not specialized too much to their own partitions for the proposed approach to work well. Furthermore, we see that the absolute performance of the sub-agents alone is not the best proxy for the final performance, based on TREC-CAR.

\section{Reformulation Examples}
\label{sec:reformulation_examples}

Table~\ref{tab:reformulation_examples} shows four reformulation examples by various methods. The proposed method (AQA-10-Sub) performs better in the first and second examples than the other methods. Note that, despite the large diversity of  reformulations, BiDAF still returns the correct answer.

In the third example, the proposed method fails to produce the right answer whereas the other methods perform well.
In the fourth example, despite the correct answer is in the set of returned answers, the aggregator fails to set a high score for it.

\begin{table}[ht]
\begin{center}
\begin{scriptsize}
\begin{tabular}{r|p{7.2cm}|p{4.4cm}}
    \toprule
    {\bf Method} & {\bf Query} & {\bf Reference / Answer from BiDAF (F1)} \\
    \midrule
Jeopardy! & The name of this drink that can be blended or on the rocks means "daisy" in Spanish\\
SearchQA  & name drink blended rocks means daisy spanish & margarita \\ 
AQA & What name drink blended rocks mean daisy spanish? & margarita tequila daisy (0.33)\\
    & \textbf{What rock drink name means daisy spanish?} & \textbf{margarita tequila daisy mentioned (0.20)}\\
    & What name drink blended rocks means daisy spanish? & margarita tequila daisy mentioned (0.20)\\
    & What rock drinks name means daisy spanish? & margarita tequila daisy mentioned (0.20)\\
    & What name drink blended rock means daisy spanish? & margarita tequila daisy mentioned (0.20)\\
AQA-10-Full & What is drink name name drink daisy daisy? me & margarita eater jun (0.33) \\
    & What name is drink spanish? & margarita eater jun (0.33) \\
    & \textbf{What is daisy blender rock daisy spanish?? daisy spanish?} & \textbf{cocktail daisy margarita spanish (0.26)} \\
    & rock name name & cocktail daisy margarita spanish (0.25)\\
    & What name drink blended st st st st st ship ship & cocktail daisy margarita spanish (0.26)\\
AQA-10-Sub & Where is name drink?? & margarita (1.0)\\
    & \textbf{What is drink blended rock?} & \textbf{margarita (1.0)}\\
    & rock definition name & margarita (1.0) \\
    & What is name drink blended rock daisy spanish 16 daisy spanish? & margarita similarity (0.5) \\
    & Nam Nam Nam Nam Nam Nam Nam drink & tequila (0.0) \\
    \midrule
Jeopardy! & A graduate of Howard University, she won the Nobel Prize for literature in 1993 & \\
SearchQA & graduate howard university , nobel prize literature 1993 & toni morrison\\ 
AQA & Nobel university of howard university? & toni morrison american novelist (0.5)\\
    & Nobel university of howard university in 1993? & toni morrison american novelist (0.5)\\
    & \textbf{Nobel graduate literature in 1993?} & \textbf{toni morrison american novelist (0.5)}\\
    & Nobel university graduate howard university 1993? & princeton (0.0) \\
    & Nobel university for howard university? & columbia (0.0)\\
AQA-10-Full & Another university start howard university starther & toni morrison american novelist (0.5)\\
    & \textbf{university howard car?} & \textbf{toni morrison american novelist (0.5)}\\
    & What is howard graduate nobel? & toni morrison american novelist (0.5)\\
    & What is howard howard university? & toni morrison american novelist (0.5)\\
    & Where graduated howard university noble literature literature? & american novelist morrison (0.16) \\
AQA-10-Sub & \textbf{Where is howard university??} & \textbf{toni morrison (1.0)}\\
    & The nobel university? & toni morrison (1.0)\\
    & What name howard howard university? & toni morrison (1.0)\\
    & This howard? & toni morrison american novelist (0.5)\\
    & 1993? & howard cornell universities (0.0) \\
   \midrule
Jeopardy! & For Bill Gates, it computes to own 2 models, the 959 and the 911, from this manufacturer\\
SearchQA & bill gates , computes 2 models , 959 911 , manufacturer & porsche\\ 
AQA & \textbf{Bill gates iin computes older models? }& \textbf{porshe (1.0)} \\ 
    & Bill gates in compute gates how old are they? & porshe (1.0) \\
    & Bill gates bill gates computes mod? & porshe (1.0) \\
    & Bill gates computes 2 models pics of 959? & porshe (1.0) \\
    & Bill gates in compute gates how old is it? & porshe (1.0) \\
AQA-10-Full & Another model start bill bette & porshe (1.0) \\
    & What is an bill gates 100 car? & porshe (1.0) \\
    & What is bill bill bill bill gates computes? & porshe (1.0) \\
    & \textbf{What is manufacturer?} & \textbf{porshe (1.0)} \\
    & bill bill gats  sa computes 2 bill gats? & porshe (1.0)\\
AQA-10-Sub & Where is bill gates manufacturer? & bill gates (0.0)\\
    & \textbf{A bill gates?} & \textbf{bill gates (0.0)} \\
    & The model? & bill gates (0.0)\\ 
    & What is bill gates model? & sports car (0.0)\\
    & What model bill gates 9 58 model 9 gates? & sports car (0.0)\\
\midrule
Jeopardy! & The first written mention of this capital's name was in a 1459 document of Vlad the Impaler\\
SearchQA & first written mention capital 's name 1459 document vlad impaler & bucharest\\ 
AQA & First film was written by 1459 vlad impaler? & bucharest castle (0.5) \\
    & First film was written by 1459 vlad impalter? & bucharest castle (0.5)\\
    & First film was written by 1459 vlad impal? & bucharest castle (0.5) \\
    & First film was written by 1459 vlad impalot? & bucharest castle (0.5)\\
    & \textbf{First film was written in 1459?} & \textbf{bucharest national capital (0.33)} \\
AQA-10-Full & What is capital vlad impaler? & bucharest (1.0) \\
    & First referred capital vlad impaler impaler? & bucharest (1.0)\\
    & capital & romania 's largest city capital (0.0)\\
    & Another name start capital & romania 's largest city capital (0.0) \\
    & \textbf{capital capital vlad car capital car capital?} & \textbf{romania 's largest city capital (0.0)}\\
AQA-10-Sub & Where is vla capital capital vlad impalers? & bucharest (1.0) \\
    & What capital vlad capital document document impaler? & bucharest (1.0)\\
    & \textbf{Another capital give capital capital} & \textbf{bulgaria , hungary , romania (0.0)} \\
    & capital? & bulgaria , hungary , romania (0.0)\\
    & The name capital name? & hungary (0.0) \\
\bottomrule
\end{tabular}
\end{scriptsize}
\end{center}
\caption{Examples for the qualitative analysis on SearchQA.
In \textbf{bold} are the reformulations and answers that had the highest scores predicted by the aggregator. We only show the top-5 reformulations of each method. For a detailed analysis of the language learned by the reformulator agents, see~\citet{buck2018analyzing}.
}
\label{tab:reformulation_examples}
\end{table}

\end{appendices}

\end{document}